\documentclass[10pt,twocolumn,letterpaper]{article}

\usepackage{cvpr}
\usepackage{times}
\usepackage{epsfig}
\usepackage{graphicx}
\usepackage{amsmath}
\usepackage{amssymb}
\usepackage{subcaption}
\usepackage{float}

\usepackage[breaklinks=true,bookmarks=false]{hyperref}
\usepackage{authblk} 

\cvprfinalcopy{} 

\def\cvprPaperID{****} 

\ifcvprfinal\fi
\begin{document}

\title{Improved Regularization of Convolutional Neural Networks with Cutout}


\author[1]{Terrance DeVries}
\author[1,2]{Graham W.~Taylor}
\affil[1]{University of Guelph}
\affil[2]{Canadian Institute for Advanced Research and Vector Institute}

\maketitle

\begin{abstract}
Convolutional neural networks are capable of learning powerful representational spaces, which are necessary for tackling complex learning tasks. However, due to the model capacity required to capture such representations, they are often susceptible to overfitting and therefore require proper regularization in order to generalize well.

In this paper, we show that the simple regularization technique of randomly masking out square regions of input during training, which we call cutout, can be used to improve the robustness and overall performance of convolutional neural networks.
Not only is this method extremely easy to implement, but we also demonstrate that it can be used in conjunction with existing forms of data augmentation and other regularizers to further improve model performance. We evaluate this method by applying it to current state-of-the-art architectures on the CIFAR-10, CIFAR-100, and SVHN datasets, yielding new state-of-the-art results of 2.56\%, 15.20\%, and 1.30\% test error respectively. 
Code available at \url{https://github.com/uoguelph-mlrg/Cutout}.

\end{abstract}

\vspace{-0.4cm} 
\section{Introduction}
In recent years deep learning has contributed to considerable advances in the field of computer vision, resulting in state-of-the-art performance in many challenging vision tasks such as object recognition~\cite{krizhevsky2012imagenet}, semantic segmentation~\cite{long2015fully}, image captioning~\cite{vinyals2015show}, and human pose estimation~\cite{toshev2014deeppose}. Much of these improvements can be attributed to the use of convolutional neural networks (CNNs)~\cite{lecun1998gradient}, which are capable of learning complex hierarchical feature representations of images. As the complexity of the task to be solved increases, the resource utilization of such models increases as well: memory footprint, parameters, operations count, inference time and power consumption~\cite{Canziani2016-si}.
Modern networks commonly contain on the order of tens to hundreds of millions of learned parameters which provide the necessary representational power for such tasks, but with the increased representational power also comes increased probability of overfitting, leading to poor generalization.

In order to combat the potential for overfitting, several different regularization techniques can be applied, such as data augmentation or the judicious addition of noise to activations, parameters, or data. In the domain of computer vision, data augmentation is almost ubiquitous due to its ease of implementation and effectiveness. Simple image transforms such as mirroring or cropping can be applied to create new training data which can be used to improve model robustness and increase accuracy~\cite{lecun1998gradient}. Large models can also be regularized by adding noise during the training process, whether it be added to the input, weights, or gradients. One of the most common uses of noise for improving model accuracy is dropout~\cite{hinton2012improving}, which stochastically drops neuron activations during training and as a result discourages the co-adaptation of feature detectors.

\begin{figure}[t]
\begin{center}
   \includegraphics[width=\linewidth, trim={0 6.015cm 0 0}, clip]{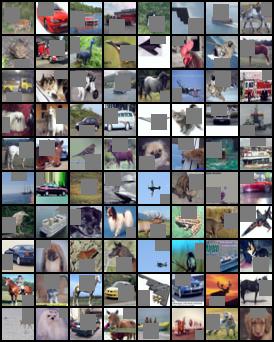}
   \caption{Cutout applied to images from the CIFAR-10 dataset.}
\label{fig:cutout_example}
\end{center}
\vspace{-0.5cm}
\end{figure}

In this work we consider applying noise in a similar fashion to dropout, but with two important distinctions. The first difference is that units are dropped out only at the input layer of a CNN, rather than in the intermediate feature layers. The second difference is that we drop out contiguous sections of inputs rather than individual pixels, as demonstrated in Figure~\ref{fig:cutout_example}. In this fashion, dropped out regions are propagated through all subsequent feature maps, producing a final representation of the image which contains no trace of the removed input, other than what can be recovered by its context. This technique encourages the network to better utilize the full context of the image, rather than relying on the presence of a small set of specific visual features. This method, which we call cutout, can be interpreted as applying a spatial prior to dropout in input space, much in the same way that convolutional neural networks leverage information about spatial structure in order to improve performance over that of feed-forward networks.


In the remainder of this paper, we introduce cutout and demonstrate that masking out contiguous sections of the input to convolutional neural networks can improve model robustness and ultimately yield better model performance. We show that this simple method works in conjunction with other current state-of-the-art techniques such as residual networks and batch normalization, and can also be combined with most regularization techniques, including standard dropout and data augmentation. Additionally, cutout can be applied during data loading in parallel with the main training task, making it effectively computationally free. To evaluate this technique we conduct tests on several popular image recognition datasets, achieving state-of-the-art results on CIFAR-10, CIFAR-100, and SVHN\@. We also achieve competitive results on  STL-10, demonstrating the usefulness of cutout for low data and higher resolution problems.

\section{Related Work}

Our work is most closely related to two common regularization techniques: data augmentation and dropout. Here we examine the use of both methods in the setting of training convolutional neural networks. We also discuss denoising auto-encoders and context encoders, which share some similarities with our work.

\subsection{Data Augmentation for Images}


Data augmentation has long been used in practice when training convolutional neural networks. When training LeNet5~\cite{lecun1998gradient} for optical character recognition, LeCun \etal{} apply various affine transforms, including horizontal and vertical translation, scaling, squeezing, and horizontal shearing to improve their model's accuracy and robustness.

In~\cite{bengio2011deep}, Bengio \etal{} demonstrate that deep architectures benefit much more from data augmentation than shallow architectures. They apply a large variety of transformations to their handwritten character dataset, including local elastic deformation, motion blur, Gaussian smoothing, Gaussian noise, salt and pepper noise, pixel permutation, and adding fake scratches and other occlusions to the images, in addition to affine transformations.

To improve the performance of AlexNet~\cite{krizhevsky2012imagenet} for the 2012 ImageNet Large Scale Visual Recognition Competition, Krizhevsky \etal{} apply image mirroring, cropping, as well as randomly adjusting colour and intensity values based on ranges determined using principal component analysis on the dataset.

Wu \etal{} take a more aggressive approach with image augmentation when training Deep Image~\cite{wu2015deep} on the ImageNet dataset. In addition to flipping and cropping they apply a wide range of colour casting, vignetting, rotation, and lens distortion (pin cushion and barrel distortion), as well as horizontal and vertical stretching.

Lemley \etal{} tackle the issue of data augmentation with a learned end-to-end approach called Smart Augmentation~\cite{lemley2017smart} instead of relying on hard-coded transformations. In this method, a neural network is trained to intelligently combine existing samples in order to generate additional data that is useful for the training process.

Of these techniques ours is closest to the occlusions applied in~\cite{bengio2011deep}, however their occlusions generally take the form of scratches, dots, or scribbles that overlay the target character, while we use zero-masking to completely obstruct an entire region.


\subsection{Dropout in Convolutional Neural Networks}

Another common regularization technique is dropout~\cite{hinton2012improving, srivastava2014dropout}, which was first introduced by Hinton \etal. Dropout is implemented by setting hidden unit activations to zero with some fixed probability during training. All activations are kept when evaluating the network, but the resulting output is scaled according to the dropout probability. This technique has the effect of approximately averaging over an exponential number of smaller sub-networks, and works well as a robust type of bagging, which discourages the co-adaptation of feature detectors within the network.

While dropout was found to be very effective at regularizing fully-connected layers, it appears to be less powerful when used with convolutional layers~\cite{tompson2015efficient}. This reduction in potency can largely be attributed to two factors. The first is that convolutional layers already have much fewer parameters than fully-connected layers, and therefore require less regularization. The second factor is that neighbouring pixels in images share much of the same information. If any of them are dropped out then the information they contain will likely still be passed on from the neighbouring pixels that are still active. For these reasons, dropout in convolutional layers simply acts to increase robustness to noisy inputs, rather than having the same model averaging effect that is observed in fully-connected layers.

In an attempt to increase the effectiveness of dropout in convolutional layers, several variations on the original dropout formula have been proposed. Tompson \etal{} introduce SpatialDropout~\cite{tompson2015efficient}, which randomly discards entire feature maps rather than individual pixels, effectively bypassing the issue of neighbouring pixels passing similar information.

Wu and Gu propose probabilistic weighted pooling~\cite{wu2015towards}, wherein activations in each pooling region are dropped with some probability. This approach is similar to applying dropout before each pooling layer, except that instead of scaling the output with respect to the dropout probability at test time, the output of each pooling function is selected to be the sum of the activations weighted by the dropout probability. The authors claim that this approach approximates averaging over an exponential number of sub-networks as dropout does.

In a more targeted approach, Park and Kwak introduce max-drop~\cite{park2016analysis}, which drops the maximal activation across feature maps or channels with some probability. While this regularization method performed better than conventional dropout on convolutional layers in some cases, they found that when used in CNNs that utilized batch normalization, both max-drop and SpatialDropout performed worse than standard dropout.

\subsection{Denoising Auto-encoders \& Context Encoders}

Denosing auto-encoders~\cite{vincent2010stacked} and context encoders~\cite{pathak2016context} both rely on self-supervised learning to elicit useful feature representations of images. These models work by corrupting input images and requiring the network to reconstruct them using the remaining pixels as context to determine how best to fill in the blanks. Specifically, denoising auto-encoders that apply Bernoulli noise randomly erase individual pixels in the input image, while context encoders erase larger spatial regions. In order to properly fill in the missing information, the auto-encoders are forced to learn how to extract useful features from the images, rather than simply learning an identity function. As context encoders are required to fill in a larger region of the image they are required to have a better understanding of the global content of the image, and therefore they learn higher-level features compared to denoising auto-encoders~\cite{pathak2016context}. These feature representations have been demonstrated to be useful for pre-training classification, detection, and semantic segmentation models.

While removing contiguous sections of the input has previously been used as an image corruption technique, like in context encoders, to our knowledge it has not previously been applied directly to the training of supervised models.

\section{Cutout}








Cutout is a simple regularization technique for convolutional neural networks that involves removing contiguous sections of input images, effectively augmenting the dataset with partially occluded versions of existing samples. This technique can be interpreted as an extension of dropout in input space, but with a spatial prior applied, much in the same way that CNNs apply a spatial prior to achieve improved performance over feed-forward networks on image data.

From the comparison between dropout and cutout, we can also draw parallels to denoising autoencoders and context encoders. While both models have the same goal, context encoders are more effective at representation learning, as they force the model to understand the content of the image in a global sense, rather than a local sense as denoising auto-encoders do.
In the same way, cutout forces models to take more of the full image context into consideration, rather than focusing on a few key visual features, which may not always be present.

One of the major differences between cutout and other dropout variants is that units are dropped at the input stage of the network rather than in the intermediate layers. This approach has the effect that visual features, including objects that are removed from the input image, are correspondingly removed from all subsequent feature maps. Other dropout variants generally consider each feature map individually, and as a result, features that are randomly removed from one feature map may still be present in others. These inconsistencies produce a noisy representation of the input image, thereby forcing the network to become more robust to noisy inputs. In this sense, cutout is much closer to data augmentation than dropout, as it is not creating noise, but instead generating images that appear novel to the network.

\subsection{Motivation}

The main motivation for cutout comes from the problem of object occlusion, which is commonly encountered in many computer vision tasks, such as object recognition, tracking, or human pose estimation. By generating new images which simulate occluded examples, we not only better prepare the model for encounters with occlusions in the real world, but the model also learns to take more of the image context into consideration when making decisions.

\begin{figure}[t]
\begin{center}
   \includegraphics[width=\linewidth, trim={0 3.6cm 0 10.8cm}, clip]{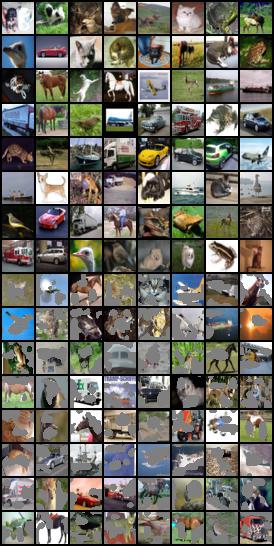}
   \caption{An early version of cutout applied to images from the CIFAR-10 dataset. This targeted approach often occludes part-level features of the image, such as heads, legs, or wheels.}
\label{fig:early_cutout}
\vspace{-0.5cm}
\end{center}
\end{figure}

We initially developed cutout as a targeted approach that specifically removed important visual features from the input of the image. This approach was similar to max-drop~\cite{park2016analysis}, in that we aimed to remove maximally activated features in order to encourage the network to consider less prominent features. To accomplish this goal, we extracted and stored the maximally activated feature map for each image in the dataset at each epoch. During the next epoch we then upsampled the saved feature maps back to the input resolution, and thresholded them at the mean feature map value to obtain a binary mask, which was finally overlaid on the original image before being passed through the CNN\@. Figure~\ref{fig:early_cutout} demonstrates this early version of cutout.

While this targeted cutout method performed well, we found that randomly removing regions of a fixed size performed just as well as the targeted approach, without requiring any manipulation of the feature maps. Due to the inherent simplicity of this alternative approach, we focus on removing fixed-size regions for all of our experiments.

\subsection{Implementation Details}
\label{details}

To implement cutout, we simply apply a fixed-size zero-mask to a random location of each input image during each epoch of training, as shown in Figure~\ref{fig:cutout_example}. Unlike dropout and its variants, we do not apply any rescaling of weights at test time. For best performance, the dataset should be normalized about zero so that modified images will not have a large effect on the expected batch statistics.

In general, we found that the size of the cutout region is a more important hyperparameter than the shape, so for simplicity, we conduct all of our experiments using a square patch as the cutout region. When cutout is applied to an image, we randomly select a pixel coordinate within the image as a center point and then place the cutout mask around that location. This method allows for the possibility that not all parts of the cutout mask are contained within the image. Interestingly, we found that allowing portions of the patches to lay outside the borders of the image (rather than constraining the entire patch to be within the image) was critical to achieving good performance. Our explanation for this phenomenon is that it is important for the model to receive some examples where a large portion of the image is visible during training. An alternative approach that achieves similar performance is to randomly apply cutout constrained within the image region, but with 50\% probability so that the network sometimes receives unmodified images.

The cutout operation can easily be applied on the CPU along with any other data augmentation steps during data loading.  By implementing this operation on the CPU in parallel with the main GPU training task, we can hide the computation and obtain performance improvements for virtually free.


\section{Experiments}

\begin{figure*}[t]
    \centering
    \begin{subfigure}[t]{0.48\textwidth}
        \includegraphics[width=\textwidth, trim={0.3cm 0 1cm 1cm}, clip]{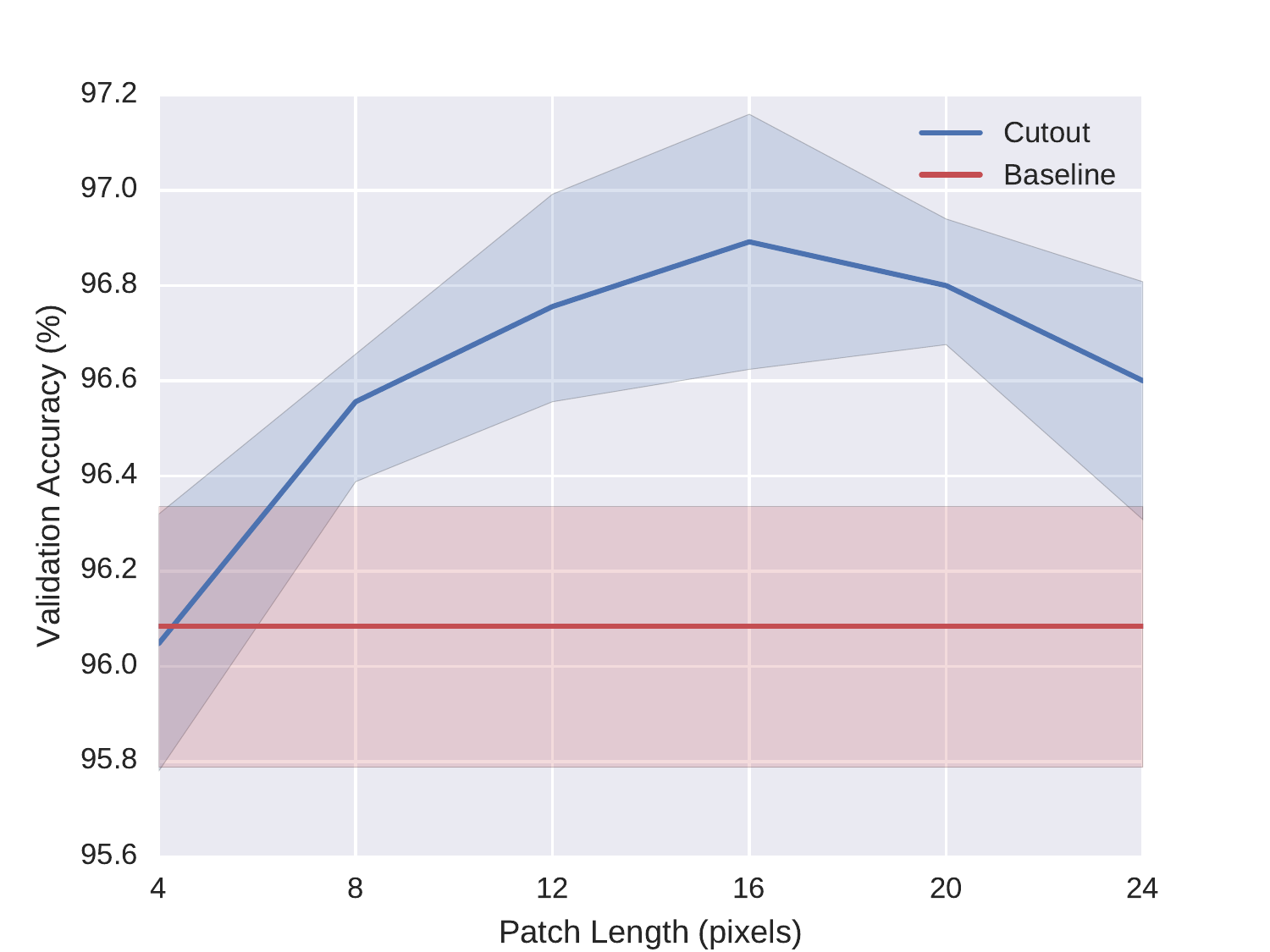}
        \caption{CIFAR-10}
\label{fig:cifar10_grid_search}
    \end{subfigure}
    \quad 
    \begin{subfigure}[t]{0.48\textwidth}
        \includegraphics[width=\textwidth, trim={0.3cm 0 1cm 1cm}, clip]{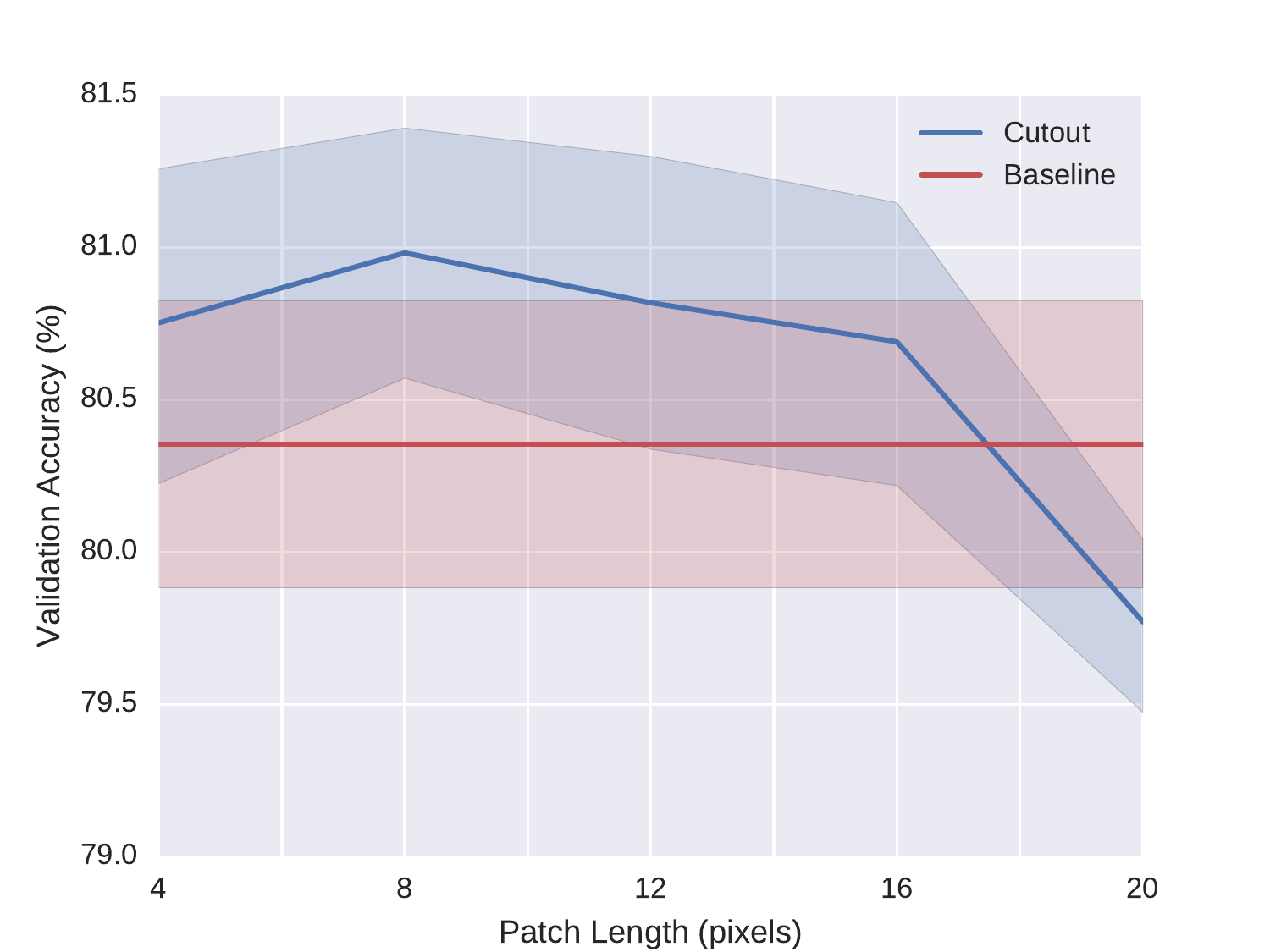}
        \caption{CIFAR-100}
\label{fig:cifar100_grid_search}
    \end{subfigure}
    \caption{Cutout patch length with respect to validation accuracy with 95\% confidence intervals (average of five runs). Tests run on CIFAR-10 and CIFAR-100 datasets using WRN-28-10 and standard data augmentation. Baseline indicates a model trained with no cutout.}\label{fig:cifar_grid_search}
\end{figure*}

\begin{table*}[t]
\centering
\begin{tabular}{l|cc|cc|c}
\hline
Method   & C10 & C10+ & C100 & C100+ & SVHN \\ \hline
ResNet18~\cite{he2016identity} &  $10.63 \pm 0.26$   &   $4.72 \pm 0.21$    &   $36.68 \pm 0.57$   &  $22.46 \pm 0.31$     &  -    \\
ResNet18 + cutout   & $9.31 \pm 0.18$ & $3.99 \pm 0.13$     &      $34.98 \pm 0.29$     & $21.96 \pm 0.24$       &   -   \\ \hline
WideResNet~\cite{zagoruyko2016wide} &    $6.97 \pm 0.22$   & $3.87 \pm 0.08$      &     $26.06 \pm 0.22$       & $18.8 \pm 0.08$        &  $1.60 \pm 0.05$  \\
WideResNet + cutout  &   $\mathbf{5.54 \pm 0.08}$   & $3.08 \pm 0.16$    &   $\mathbf{23.94 \pm 0.15}$ & $18.41 \pm 0.27$       &   $\mathbf{1.30 \pm 0.03}$  \\\hline
Shake-shake regularization~\cite{gastaldi2017shake} &  -   & $2.86$ &  -  & $15.85$   &   - \\
Shake-shake regularization + cutout  &  -   & $\mathbf{2.56 \pm 0.07}$ &  -  &  $\mathbf{15.20 \pm 0.21}$   &   - \\ \hline
\end{tabular}
\caption{Test error rates (\%) on CIFAR (C10, C100) and SVHN datasets. ``+'' indicates standard data augmentation (mirror + crop). Results averaged over five runs, with the exception of shake-shake regularization which only had three runs each. Baseline shake-shake regularization results taken from~\cite{gastaldi2017shake}.}
\label{table:cifar_svhn_results}
\end{table*}

To evaluate the performance of cutout, we apply it to a variety of natural image recognition datasets: CIFAR-10, CIFAR-100, SVHN, and STL-10.

\subsection{CIFAR-10 and CIFAR-100}
\label{sec:cifar10_and_cifar100}

Both of the CIFAR datasets~\cite{krizhevsky2009learning} consist of 60,000 colour images of size $32 \times 32$ pixels. CIFAR-10 has 10 distinct classes, such as cat, dog, car, and boat. CIFAR-100 contains 100 classes, but requires much more fine-grained recognition compared to CIFAR-10 as some classes are very visually similar. For example, it contains five different classes of trees: maple, oak, palm, pine, and willow. Each dataset is split into a training set with 50,000 images and a test set with 10,000 images.

Both datasets were normalized using per-channel mean and standard deviation. When required, we apply the standard data augmentation scheme for these datasets~\cite{he2016identity}. Images are first zero-padded with 4 pixels on each side to obtain a $40 \times 40$ pixel image, then a $32 \times 32$ crop is randomly extracted. Images are also randomly mirrored horizontally with 50\% probability.

To evaluate cutout on the CIFAR datasets, we train models using two modern architectures: a deep residual network~\cite{he2016identity} with a depth of 18 (ResNet18), and a wide residual network~\cite{zagoruyko2016wide} with a depth of 28, a widening factor of 10, and dropout with a drop probability of \textit{p} = 0.3 in the convolutional layers (WRN-28-10). For both of these experiments, we use the same training procedure as specified in~\cite{zagoruyko2016wide}. That is, we train for 200 epochs with batches of 128 images using SGD, Nesterov momentum of 0.9, and weight decay of 5e-4. The learning rate is initially set to 0.1, but is scheduled to decrease by a factor of 5x after each of the 60th, 120th, and 160th epochs. We also apply cutout to shake-shake regularization models~\cite{gastaldi2017shake} that currently achieve state-of-the-art performance on the CIFAR datasets, specifically a $26\ 2 \times 96$d ``Shake-Shake-Image'' ResNet for CIFAR-10 and a $29\ 2 \times 4 \times 64$d ``Shake-Even-Image'' ResNeXt for CIFAR-100. For our tests, we use the original code and training settings provided by the author of~\cite{gastaldi2017shake}, with the only change being the addition of cutout.

To find the best parameters for cutout we isolate 10\% of the training set to use as a validation set and train on the remaining images. As our cutout shape is square, we perform a grid search over the side length parameter to find the optimal size. We find that model accuracy follows a parabolic trend, increasing proportionally to the cutout size until an optimal point, after which accuracy again decreases and eventually drops below that of the baseline model. This behaviour can be observed in Figure~\ref{fig:cifar10_grid_search} and~\ref{fig:cifar100_grid_search}, which depict the grid searches conducted on CIFAR-10 and CIFAR-100 respectively. Based on these validation results we select a cutout size of $16 \times 16$ pixels to use on CIFAR-10 and a cutout size of $8 \times 8$ pixels for CIFAR-100 when training on the full datasets. Interestingly, it appears that as the number of classes increases, the optimal cutout size decreases. This makes sense, as when more fine-grained detection is required then the context of the image will be less useful for identifying the category. Instead, smaller and more nuanced details are important.

As shown in Table~\ref{table:cifar_svhn_results}, the addition of cutout to the ResNet18 and WRN-28-10 models increased their accuracy on CIFAR-10 and CIFAR-100 by between 0.4 to 2.0 percentage points. We draw attention to the fact that cutout yields these performance improvements even when applied to complex models that already utilize batch normalization, dropout, and data augmentation. Adding cutout to the current state-of-the-art shake-shake regularization models improves performance by 0.3 and 0.6 percentage points on CIFAR-10 and CIFAR-100 respectively, yielding new state-of-the-art results of 2.56\% and 15.20\% test error.

\subsection{SVHN}
The Street View House Numbers (SVHN) dataset~\cite{netzer2011reading} contains a total of 630,420 colour images with a resolution of $32 \times 32$ pixels. Each image is centered about a number from one to ten, which needs to be identified. The official dataset split contains 73,257 training images and 26,032 test images, but there are also 531,131 additional training images available. Following standard procedure for this dataset~\cite{zagoruyko2016wide}, we use both available training sets when training our models, and do not apply any data augmentation. All images are normalized using per-channel mean and standard deviation.

To evalute cutout on the SVHN dataset we apply it to a WideResNet with a depth of 16, a widening factor of 8, and dropout on the convolutional layers with a dropout rate of \textit{p} = 0.4 (WRN-16-8). This particular configuration currently holds state-of-the-art performance on the SVHN dataset with a test error of 1.54\%~\cite{zagoruyko2016wide}. We repeat the same training procedure as specified in~\cite{zagoruyko2016wide} by training for 160 epochs with batches of 128 images. The network is optimized using SGD with Nesterov momentum of 0.9 and weight decay of 5e-4. The learning rate is initially set to 0.01, but is reduced by a factor of 10x after the 80th and 120th epochs. The one change we do make to the original training procedure (for both baseline and cutout) is to normalize the data so that it is compatible with cutout (see \S~\ref{details}). The original implementation scales data to lie between 0 and 1.

To find the optimal size for the cutout region we conduct a grid search using 10\% of the training set for validation and ultimately select a cutout size of $20 \times 20$ pixels. While this may seem like a large portion of the image to remove, it is important to remember that the cutout patches are not constrained to lie fully within the bounds of the image.

\begin{figure*}[htp]
\begin{minipage}{0.32\textwidth}
 \includegraphics[width=\linewidth, trim={1.5cm 0cm 1cm 0cm}]{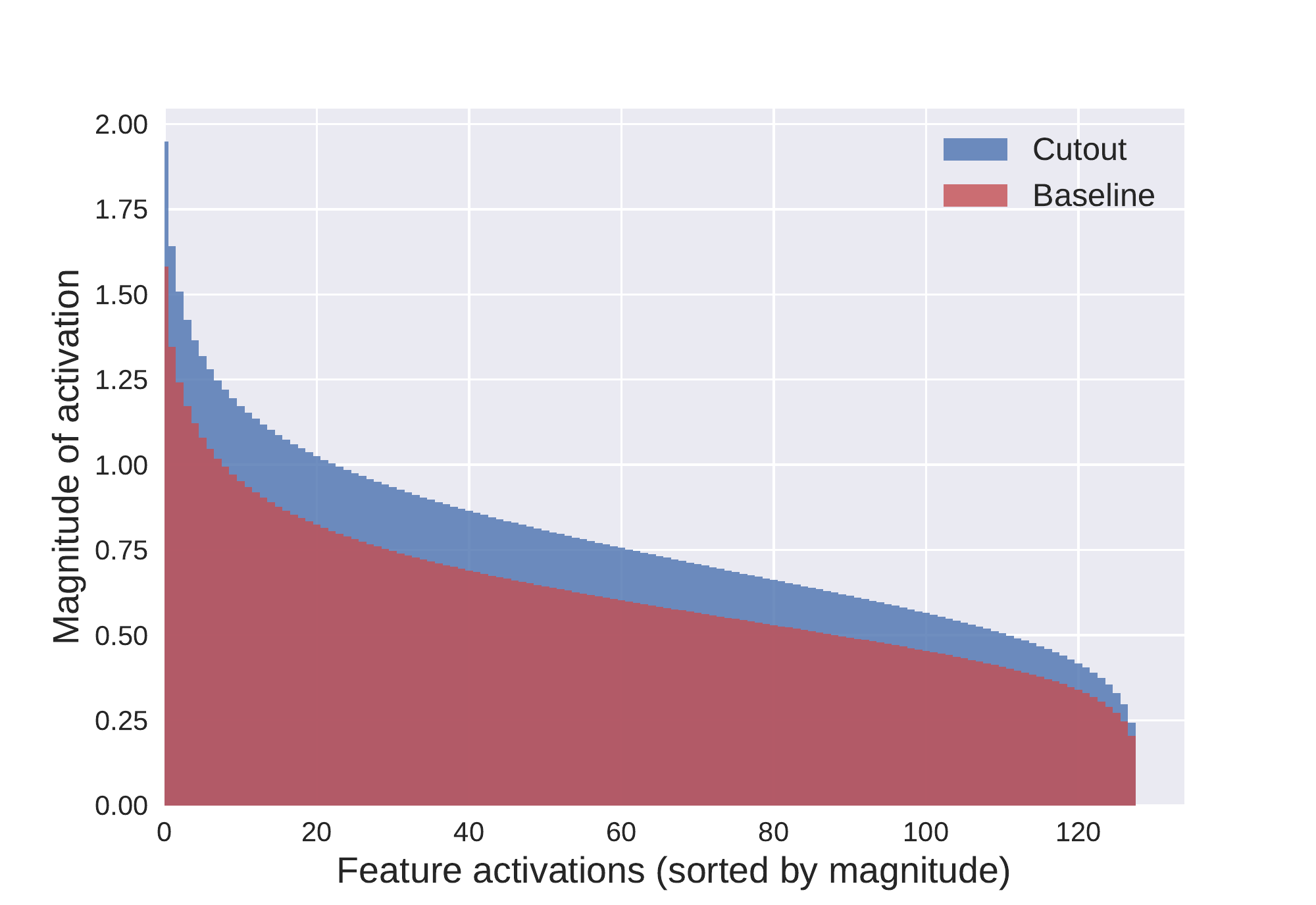}
 \subcaption{2nd Residual Block}
\end{minipage}
\begin{minipage}{0.32\textwidth}
 \includegraphics[width=\linewidth, trim={1.5cm 0cm 1cm 0cm}]{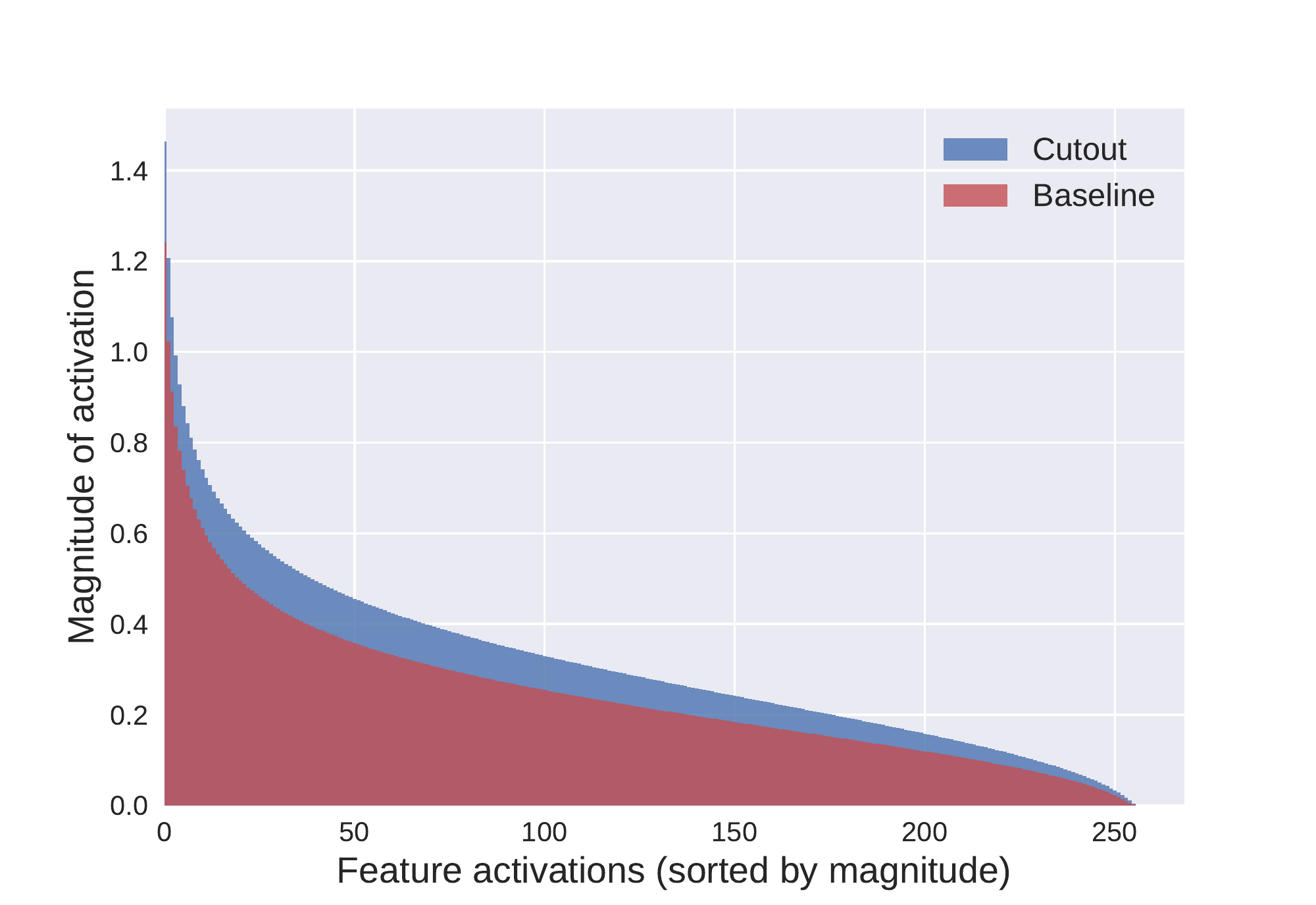}
 \subcaption{3rd Residual Block}
\end{minipage}
\begin{minipage}{0.32\textwidth}
 \includegraphics[width=\linewidth, trim={1.5cm 0cm 1cm 0cm}]{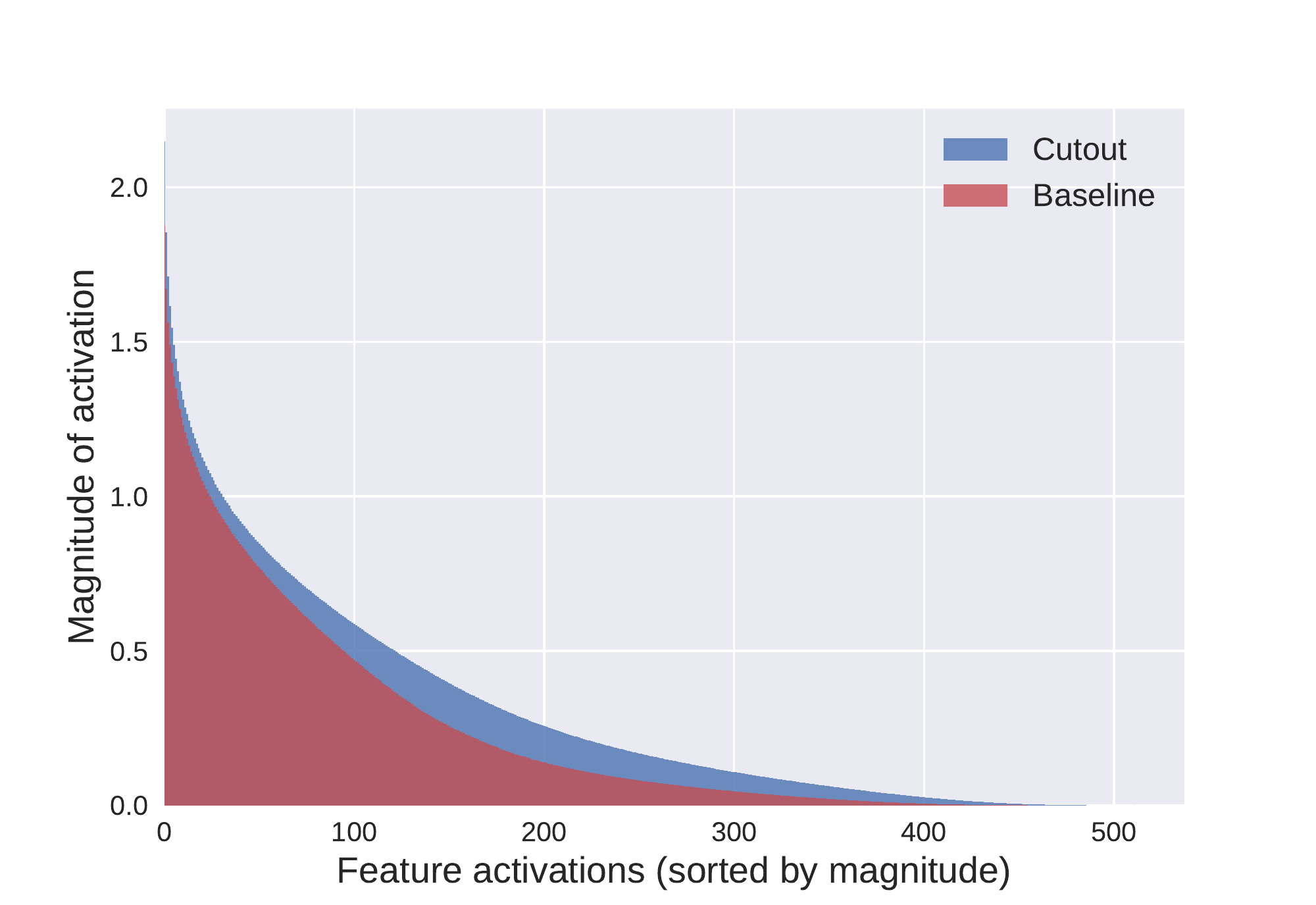}
 \subcaption{4th Residual Block}
\end{minipage}
\caption{Magnitude of feature activations, sorted by descending value, and averaged over all test samples. A standard ResNet18 is compared with a ResNet18 trained with cutout at three different depths.}
\label{fig:mean_act_hist}
\end{figure*}

Using these settings we train the WRN-16-8 and observe an average reduction in test error of 0.3 percentage points, resulting in a new state-of-the-art performance of 1.30\% test error, as shown in Table~\ref{table:cifar_svhn_results}.

\subsection{STL-10}
The STL-10 dataset~\cite{coates2011analysis} consists of a total of 113,000 colour images with a resolution of $96 \times 96$ pixels. The training set only contains 5,000 images while the test set consists of 8,000 images. All training and test set images belong to one of ten classes, such as airplane, bird, or horse. The remainder of the dataset is composed of 100,000 unlabeled images belonging to the target ten classes, plus additional but visually similar classes. While the main purpose of the STL-10 dataset is to test semi-supervised learning algorithms, we use it to observe how cutout performs when applied to higher resolution images in a low data setting. For this reason, we discard the unlabeled portion of the dataset and only use the labeled training set.

The dataset was normalized by subtracting the per-channel mean and dividing by the per-channel standard deviation. Simple data augmentation was also applied in a similar fashion to the CIFAR datasets. Specifically, images were zero-padded with 12 pixels on each side and then a $96 \times 96$ crop was randomly extracted. Mirroring horizontally was also applied with 50\% probability.

To evaluate the performance of cutout on the STL-10 dataset we use a WideResNet with a depth of 16, a widening factor of 8, and dropout with a drop rate of \textit{p} = 0.3 in the convolutional layers. We train the model for 1000 epochs with batches of 128 images using SGD with Nesterov momentum of 0.9 and weight decay of 5e-4. The learning rate is initially set to 0.1 but is reduced by a factor of 5x after the 300th, 400th, 600th, and 800th epochs.

We perform a grid search over the cutout size parameter using 10\% of the training images as a validation set and select a square size of $24 \times 24$ pixels for the no data-augmentation case and $32 \times 32$ pixels for training STL-10 with data augmentation. Training the model using these values yields a reduction in test error of 2.7 percentage points in the no data augmentation case, and 1.5 percentage points when also using data augmentation, as shown in Table~\ref{table:stl10}.

\begin{table}[h]
\centering
\begin{tabular}{l|c|c c}
\hline
\textbf{Model}      & \textbf{STL10}             & \textbf{STL10+}             \\ \hline
WideResNet          & $23.48 \pm 0.68$          & $14.21 \pm 0.29$            \\
WideResNet + cutout & $\mathbf{20.77 \pm 0.38}$          & $\mathbf{12.74 \pm 0.23}$  \\ \hline 
\end{tabular}
\caption{Test error rates on STL-10 dataset. ``+'' indicates standard data augmentation (mirror + crop). Results averaged over five runs on full training set.}
\label{table:stl10}
\end{table}

\subsection{Analysis of Cutout's Effect on Activations}
\label{analysis}

In order to better understand the effect of cutout, we compare the average magnitude of feature activations in a ResNet18 when trained with and without cutout on CIFAR-10. The models were trained with data augmentation using the same settings as defined in Section~\ref{sec:cifar10_and_cifar100}, achieving scores of 3.89\% and 4.94\% test error respectively.

In Figure~\ref{fig:mean_act_hist}, we sort the activations within each layer by ascending magnitude, averaged over all samples in the test set. We observe that the shallow layers of the network experience a general increase in activation strength, while in deeper layers, we see more activations in the tail end of the distribution. The latter observation illustrates that cutout is indeed encouraging the network to take into account a wider variety of features when making predictions, rather than relying on the presence of a smaller number of features. Figure~\ref{fig:act_hist} demonstrates similar observations for individual samples, where the effects of cutout are more pronounced.

\section{Conclusion}
Cutout was originally conceived as a targeted method for removing visual
features with high activations in later layers of a CNN\@. Our motivation was to
encourage the network to focus more on complimentary and less prominent features, in
order to generalize to situations like occlusion. However, we discovered that
the conceptually and computationally simpler approach of randomly masking square
sections of the image performed equivalently in the experiments we conducted.
Importantly, this simple regularizer proved to be complementary to
existing forms of data augmentation and regularization. Applied to modern
architectures, such as wide residual networks or shake-shake regularization models,
it achieves state-of-the-art performance on the CIFAR-10,
CIFAR-100, and SVHN vision benchmarks. So why hasn't it been reported or
analyzed to date?
One reason could be the fact that using a combination of corrupted and clean images appears to be important for its success.
Future work will return to
our original investigation of visual feature removal informed by activations.

\newpage

\section*{Acknowledgements}
{\small
The authors thank Daniel Jiwoong Im for feedback on the paper and for suggesting
the analysis in \S~\ref{analysis}. The authors also thank NVIDIA for the
donation of a Titan X GPU.}

{\small
\bibliographystyle{ieee}
\bibliography{egbib}
}

\clearpage

\appendix
\onecolumn
\section{Supplementary Materials}

\begin{figure}[h]
\centering
\begin{minipage}{0.32\textwidth}
 \includegraphics[width=\linewidth, trim={1.5cm 0cm 1cm 0cm}]{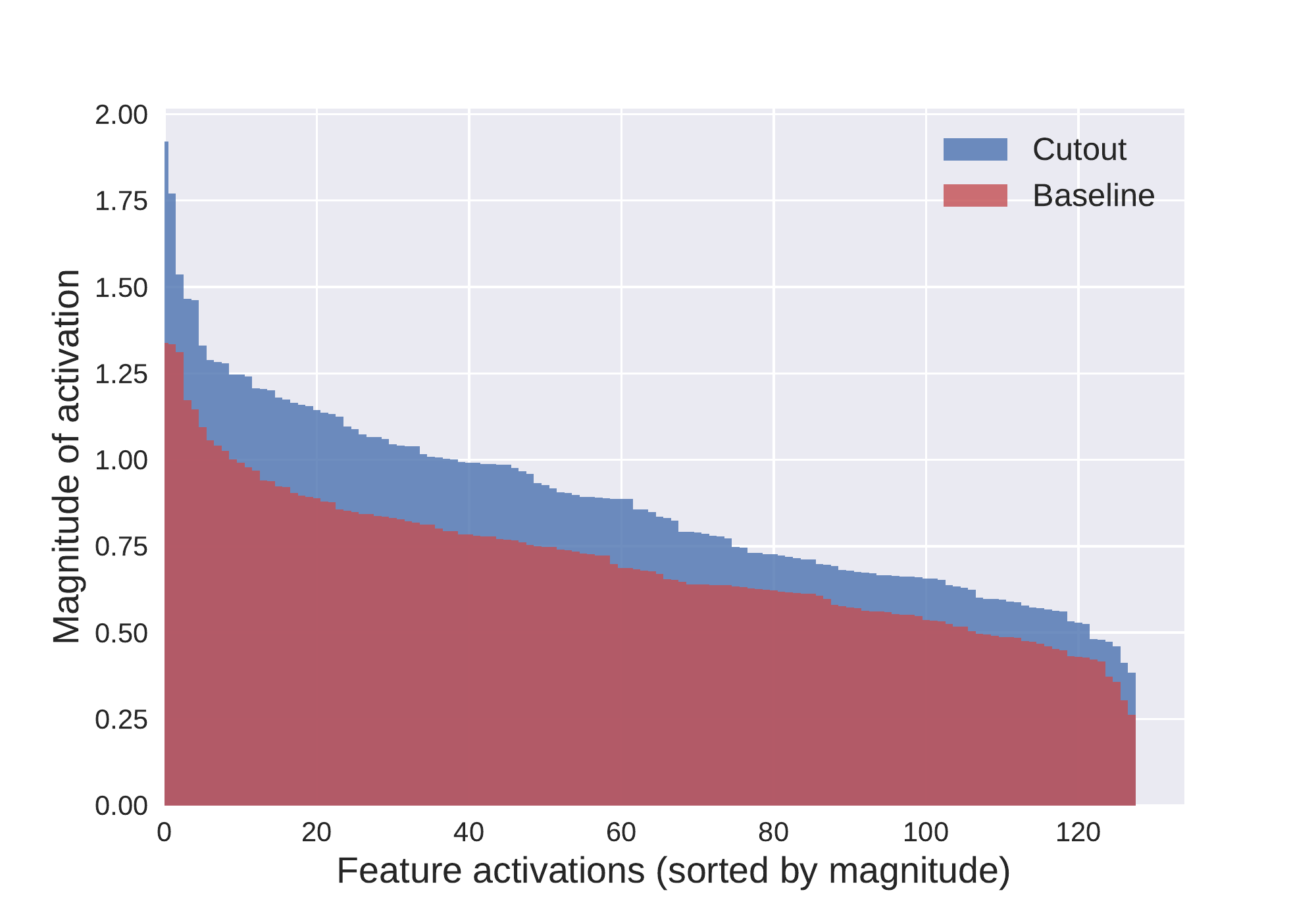}
 \subcaption{2nd Residual Block}
\end{minipage}
\begin{minipage}{0.32\textwidth}
 \includegraphics[width=\linewidth, trim={1.5cm 0cm 1cm 0cm}]{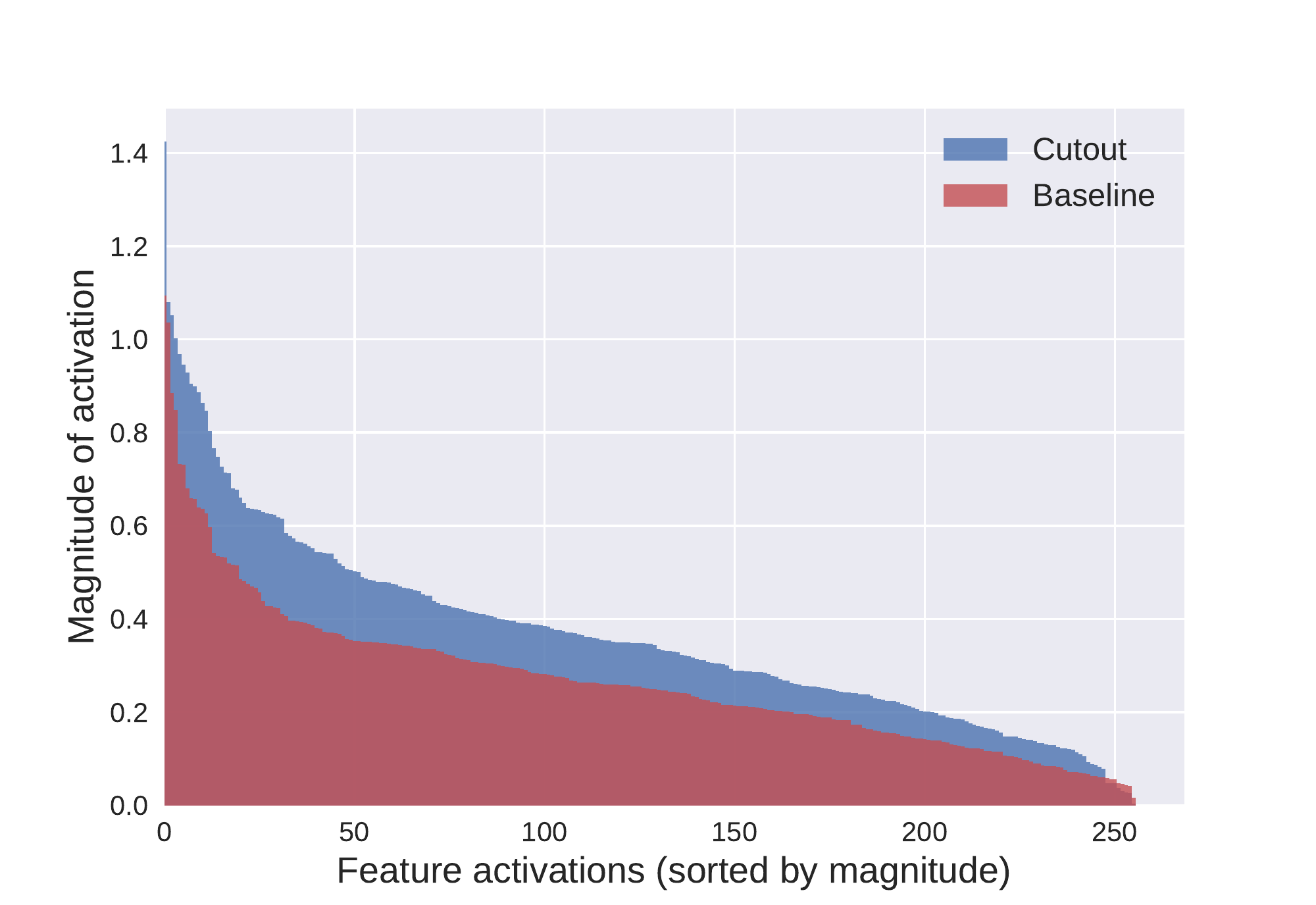}
 \subcaption{3rd Residual Block}
\end{minipage}
\begin{minipage}{0.32\textwidth}
 \includegraphics[width=\linewidth, trim={1.5cm 0cm 1cm 0cm}]{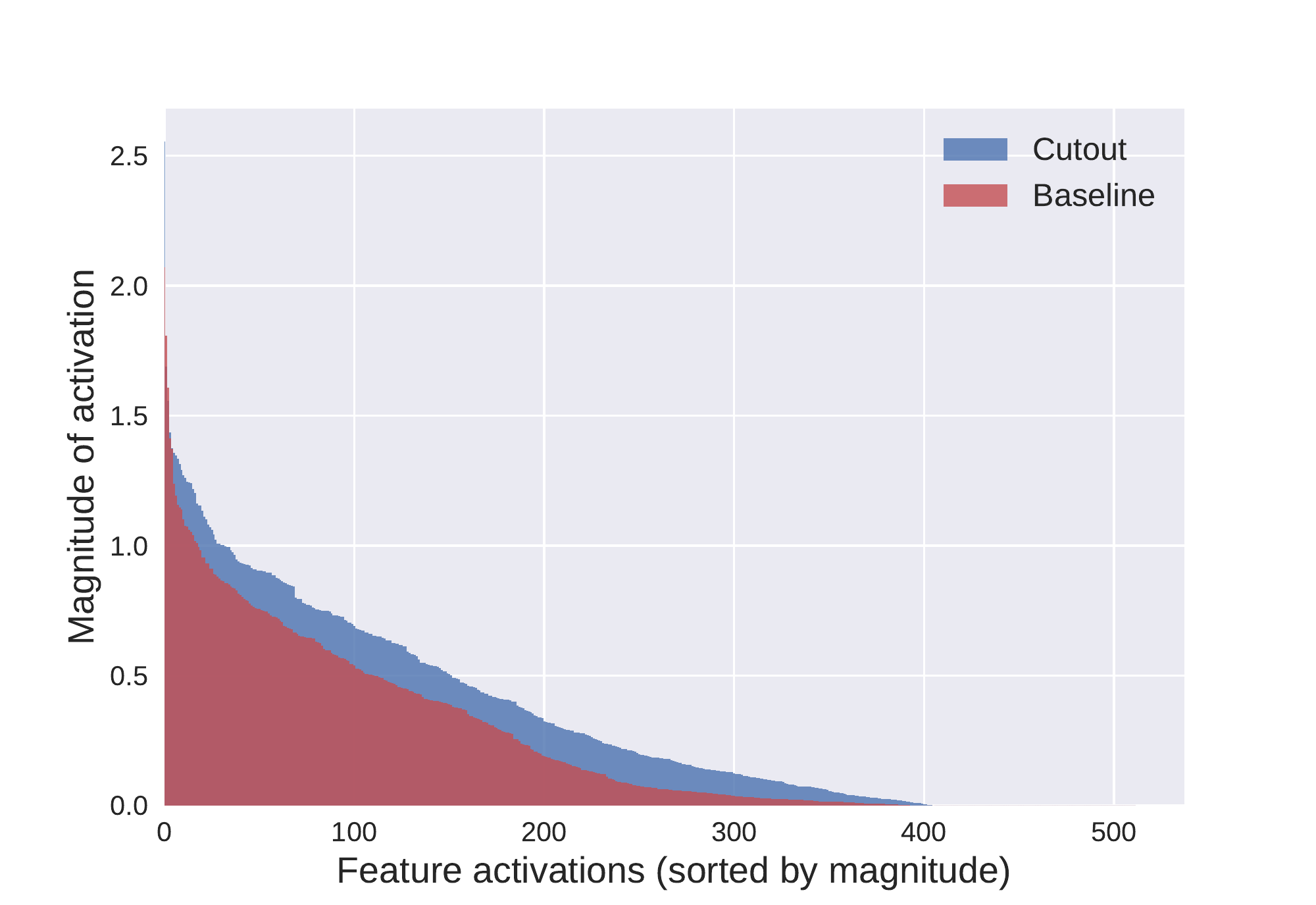}
 \subcaption{4th Residual Block}
\end{minipage}\\
\begin{minipage}{0.32\textwidth}
 \includegraphics[width=\linewidth, trim={1.5cm 0cm 1cm 0cm}]{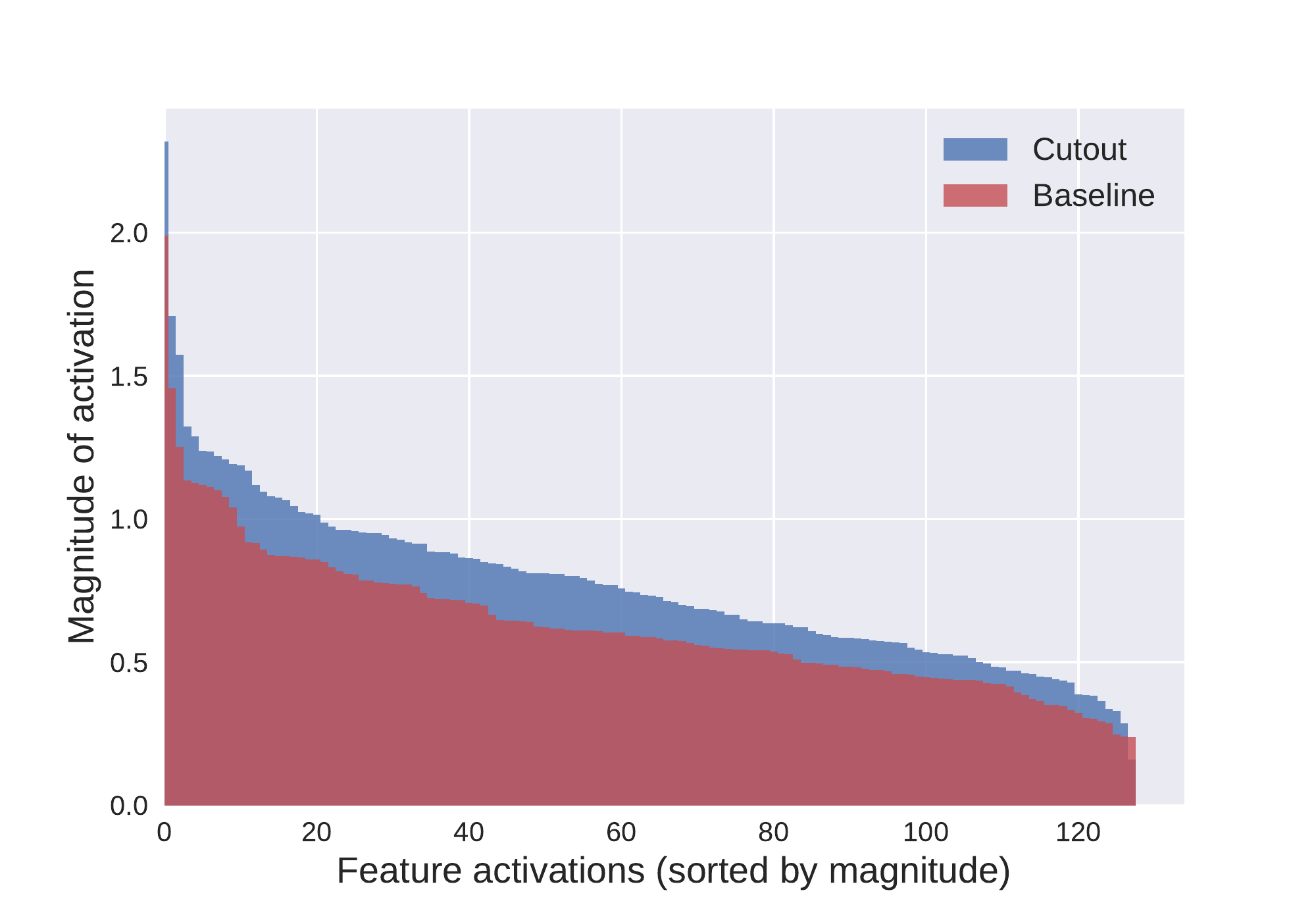}
 \subcaption{2nd Residual Block}
\end{minipage}
\begin{minipage}{0.32\textwidth}
 \includegraphics[width=\linewidth, trim={1.5cm 0cm 1cm 0cm}]{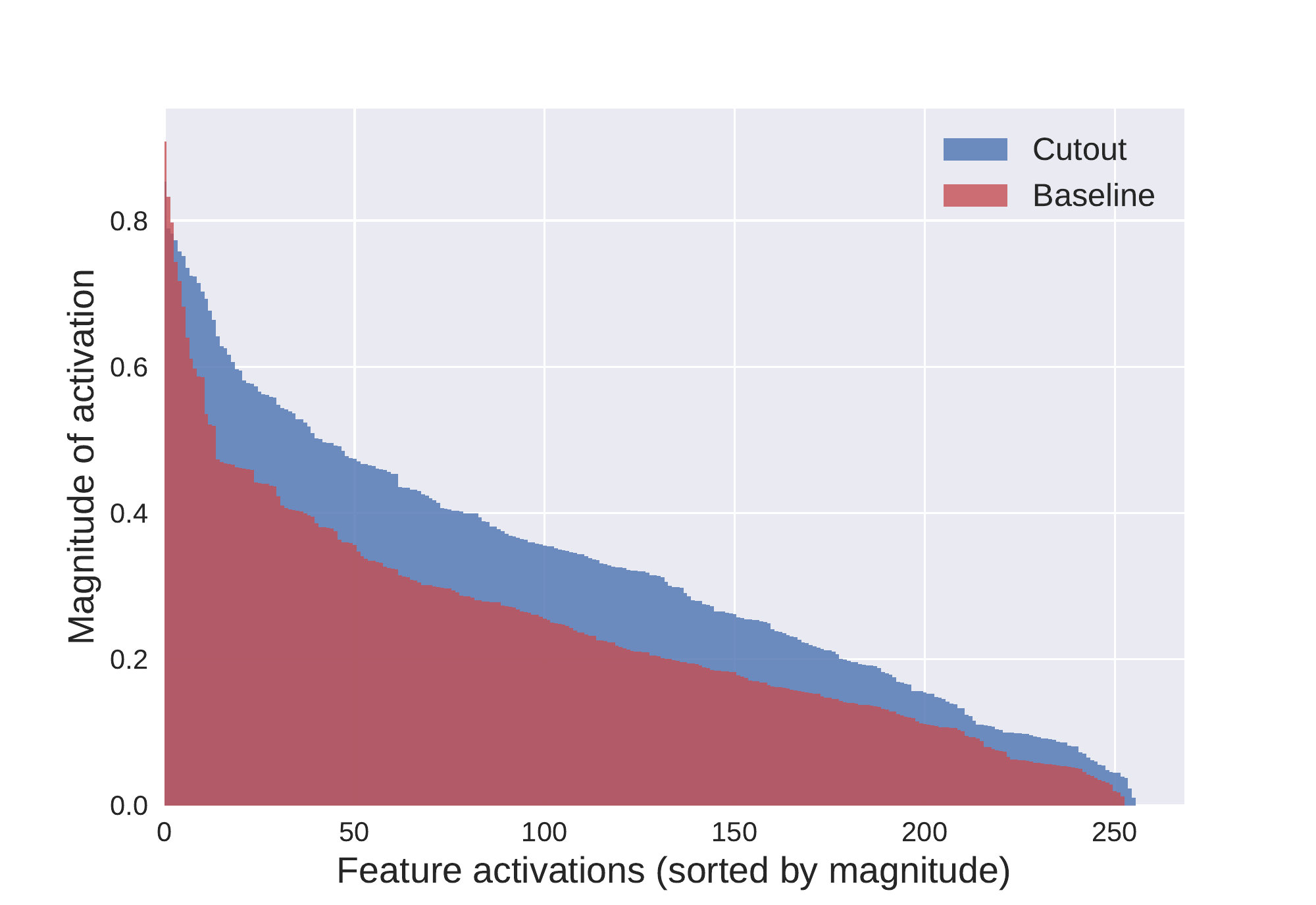}
 \subcaption{3rd Residual Block}
\end{minipage}
\begin{minipage}{0.32\textwidth}
 \includegraphics[width=\linewidth, trim={1.5cm 0cm 1cm 0cm}]{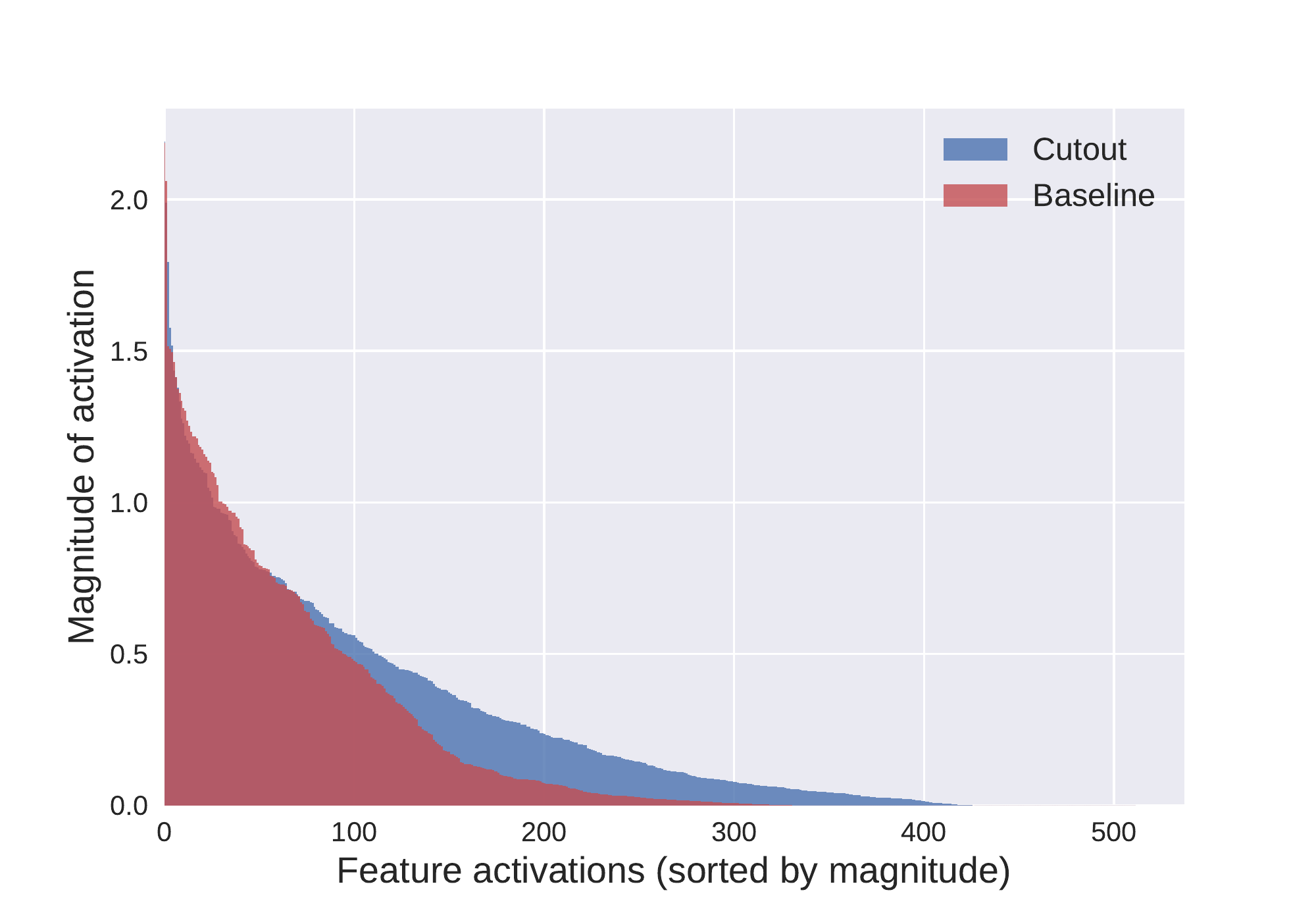}
 \subcaption{4th Residual Block}
\end{minipage}\\
\begin{minipage}{0.32\textwidth}
 \includegraphics[width=\linewidth, trim={1.5cm 0cm 1cm 0cm}]{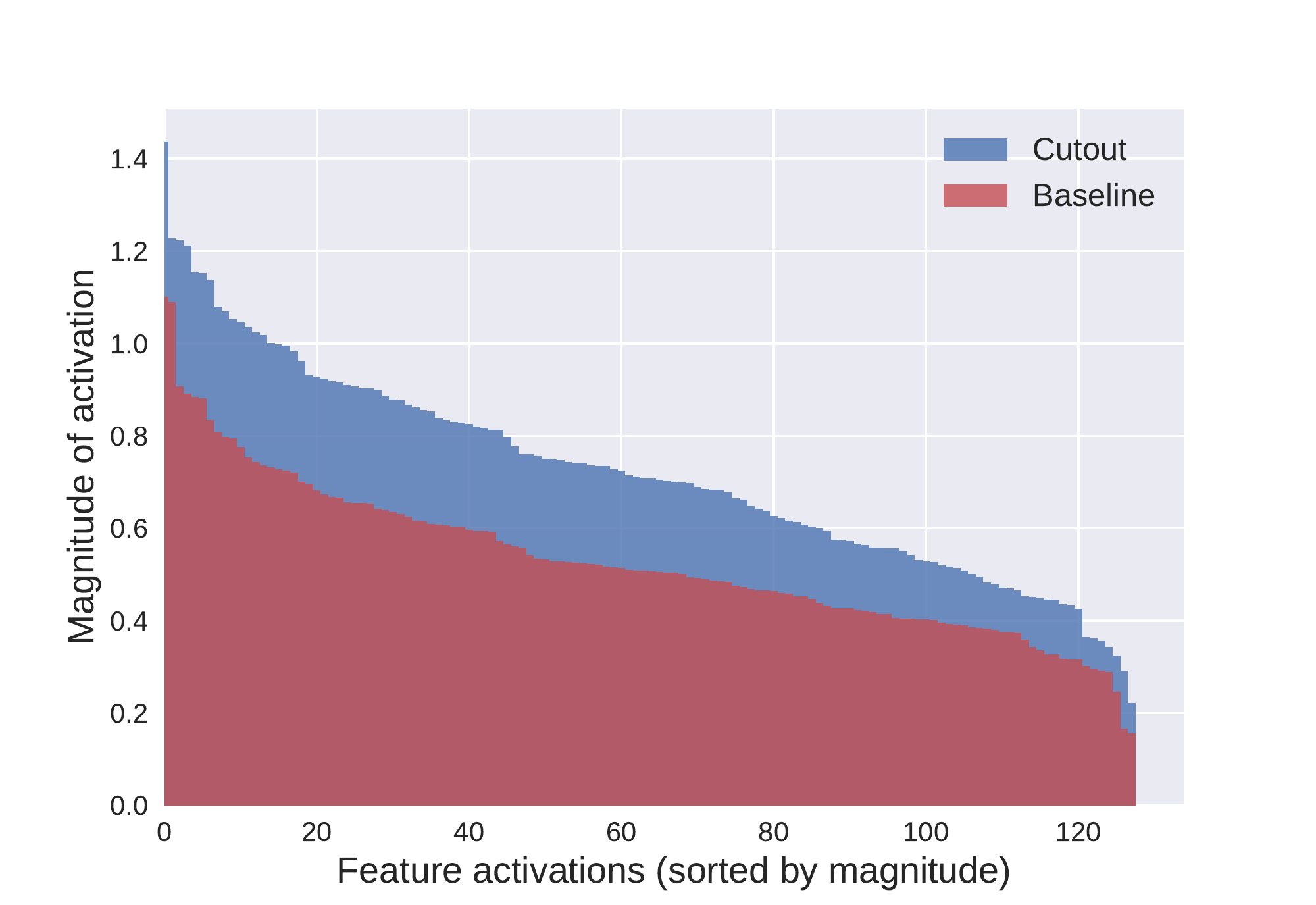}
 \subcaption{2nd Residual Block}
\end{minipage}
\begin{minipage}{0.32\textwidth}
 \includegraphics[width=\linewidth, trim={1.5cm 0cm 1cm 0cm}]{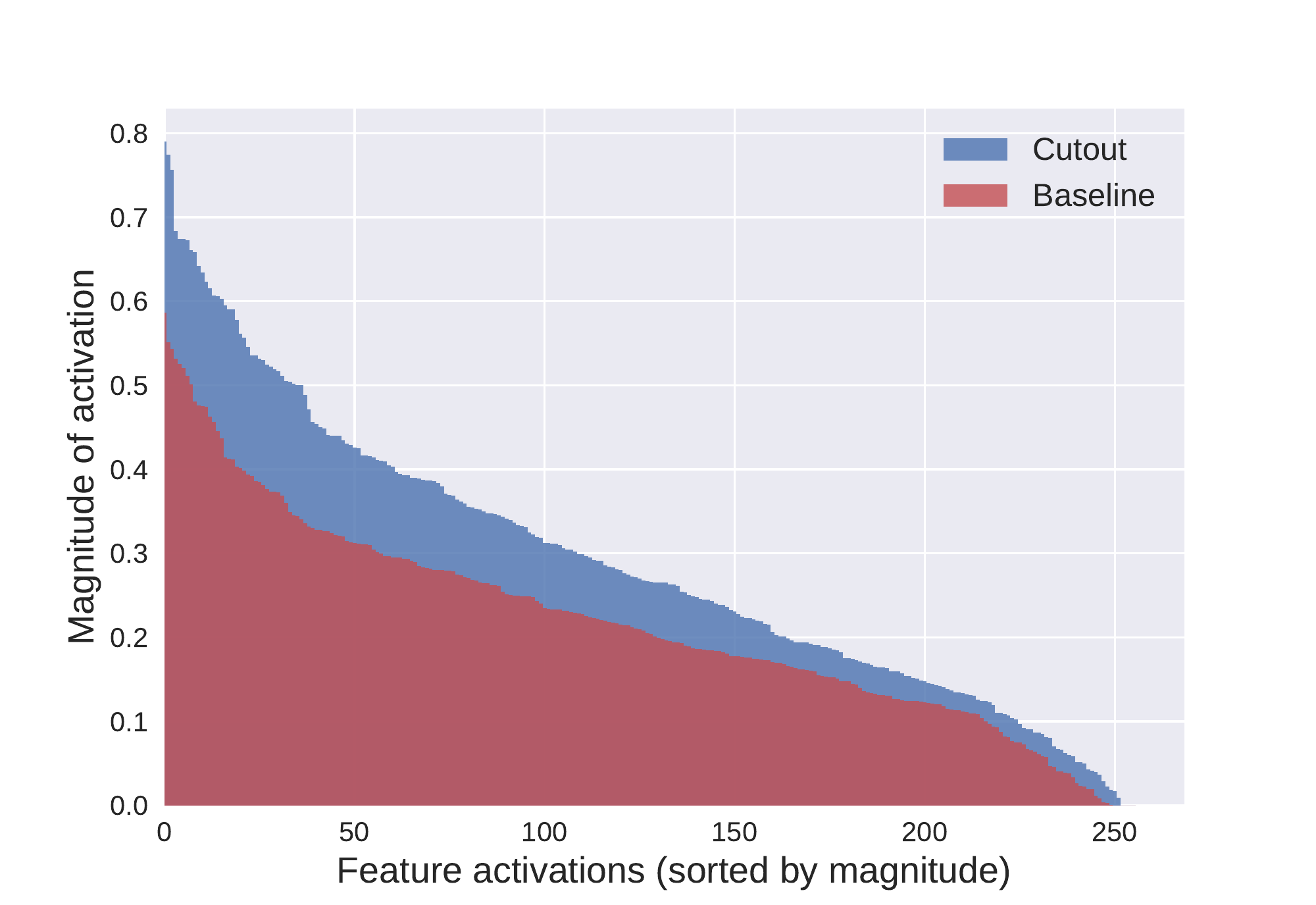}
 \subcaption{3rd Residual Block}
\end{minipage}
\begin{minipage}{0.32\textwidth}
 \includegraphics[width=\linewidth, trim={1.5cm 0cm 1cm 0cm}]{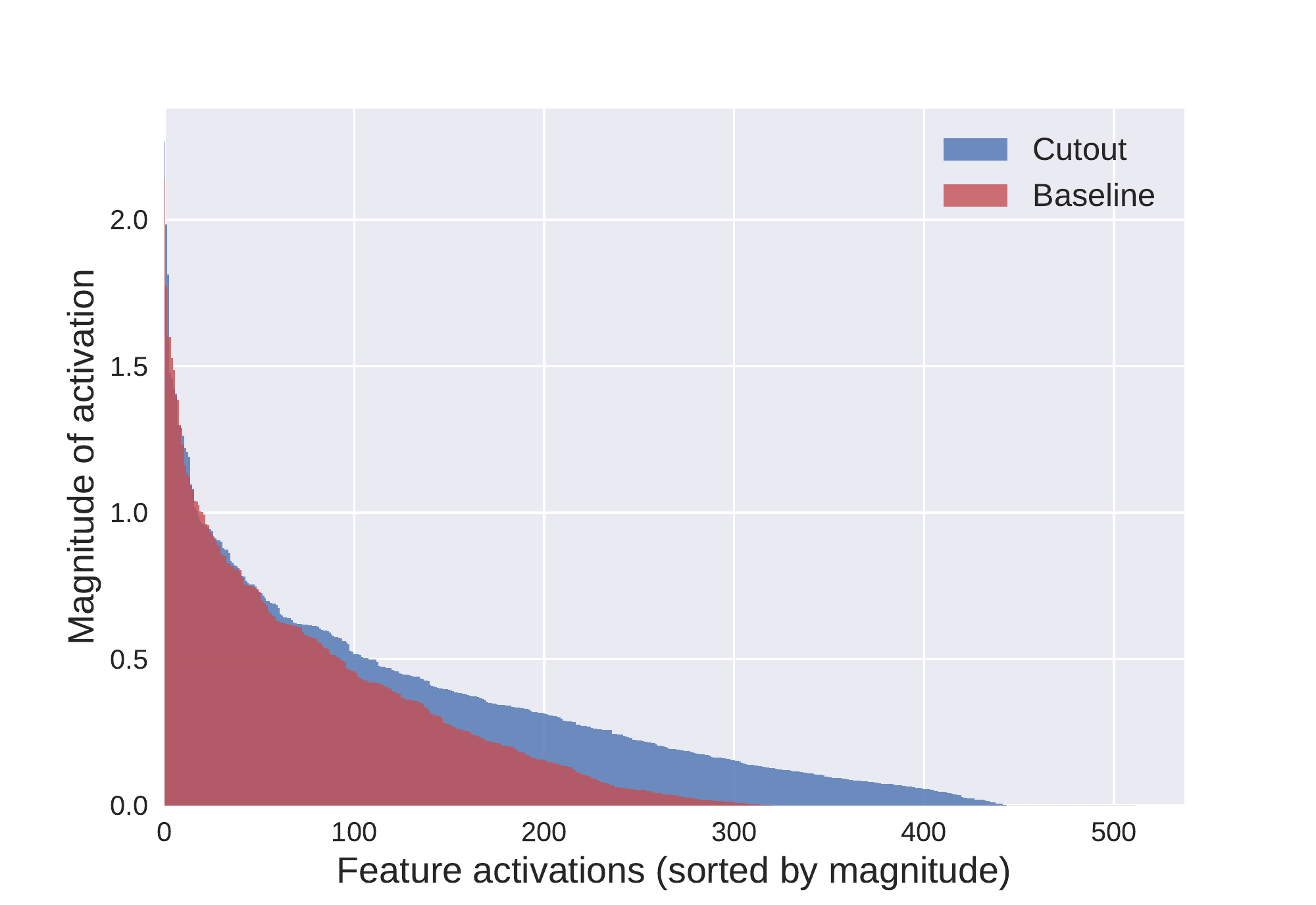}
 \subcaption{4th Residual Block}
\end{minipage}\\
\begin{minipage}{0.32\textwidth}
 \includegraphics[width=\linewidth, trim={1.5cm 0cm 1cm 0cm}]{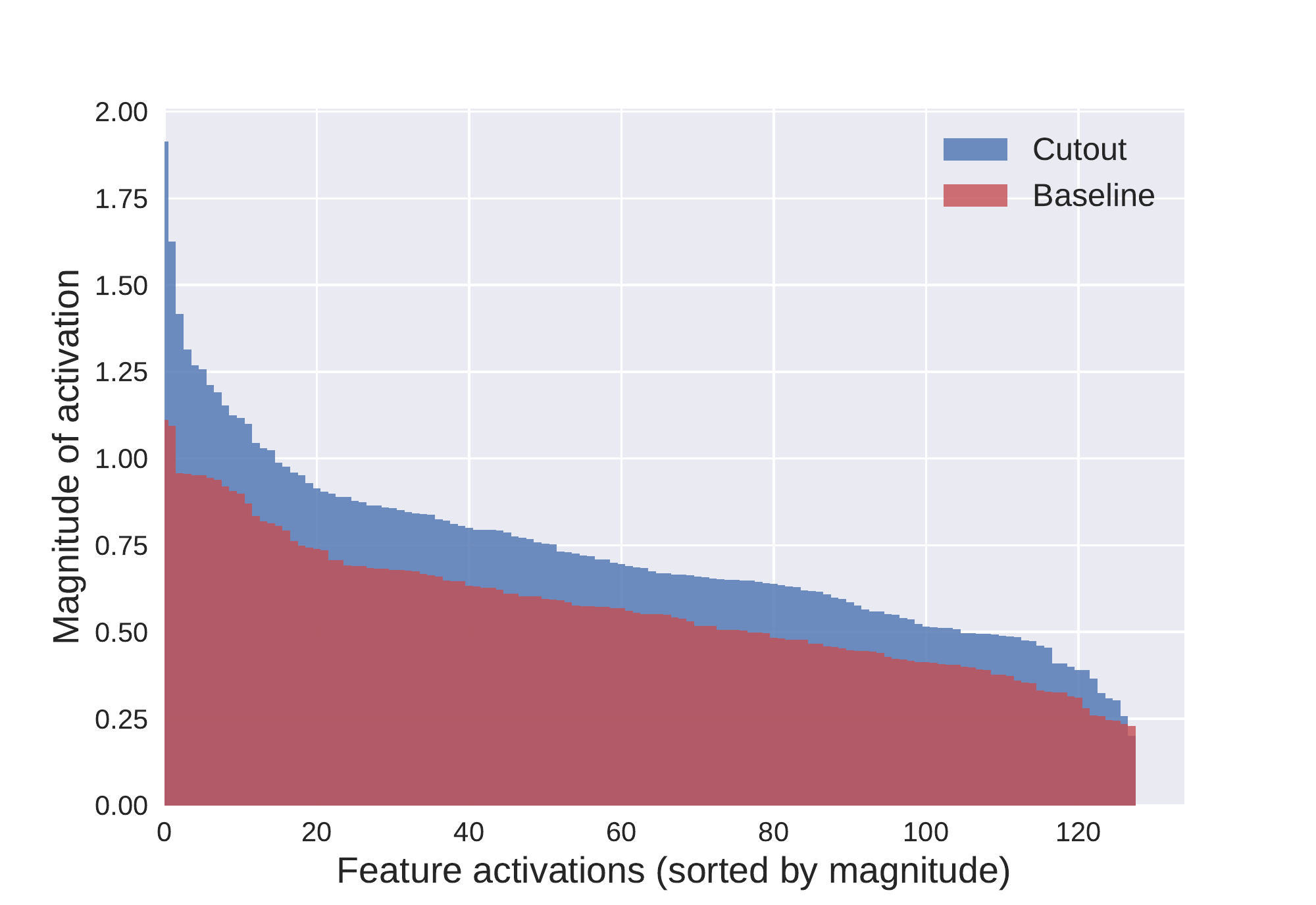}
 \subcaption{2nd Residual Block}
\end{minipage}
\begin{minipage}{0.32\textwidth}
 \includegraphics[width=\linewidth, trim={1.5cm 0cm 1cm 0cm}]{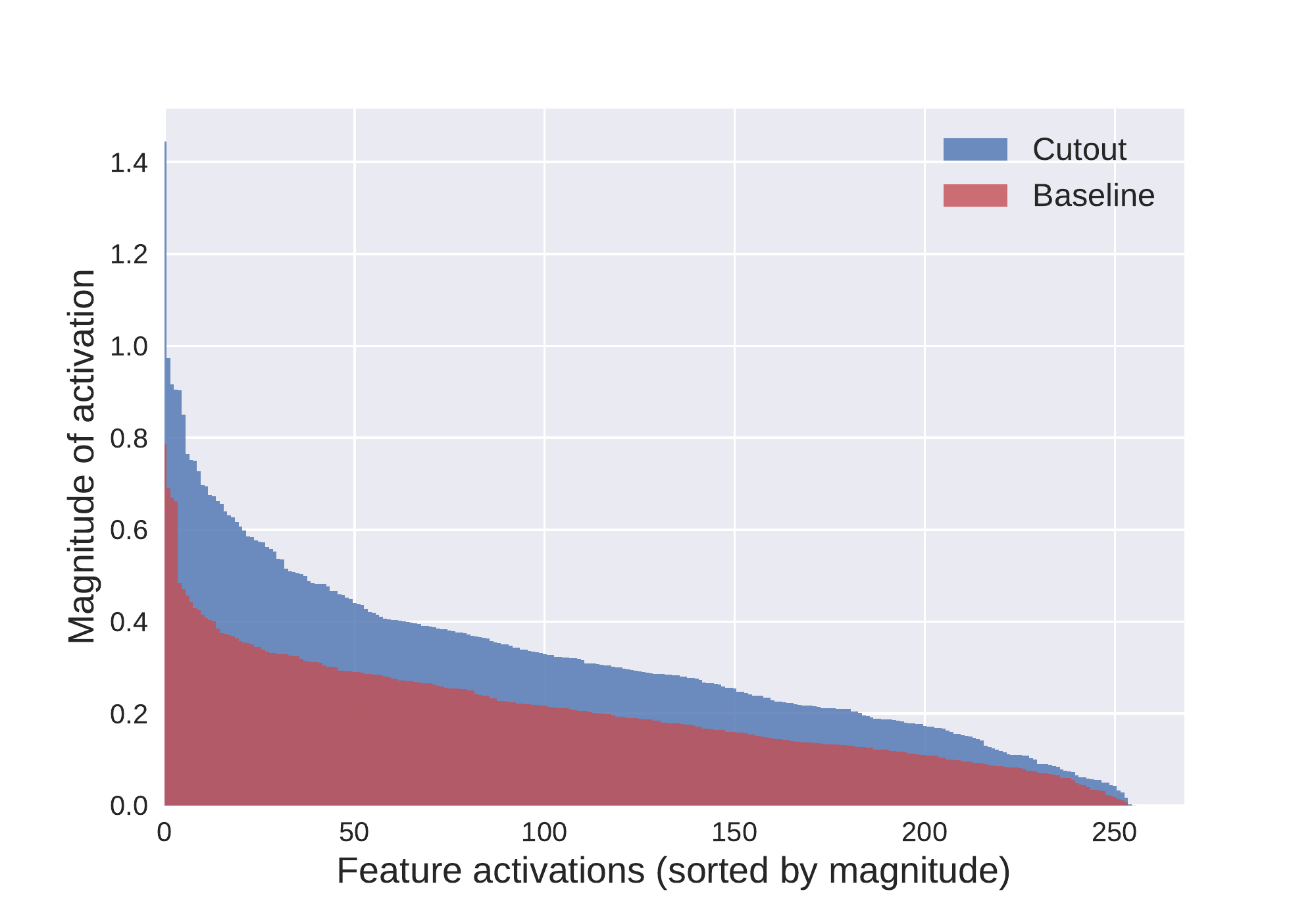}
 \subcaption{3rd Residual Block}
\end{minipage}
\begin{minipage}{0.32\textwidth}
 \includegraphics[width=\linewidth, trim={1.5cm 0cm 1cm 0cm}]{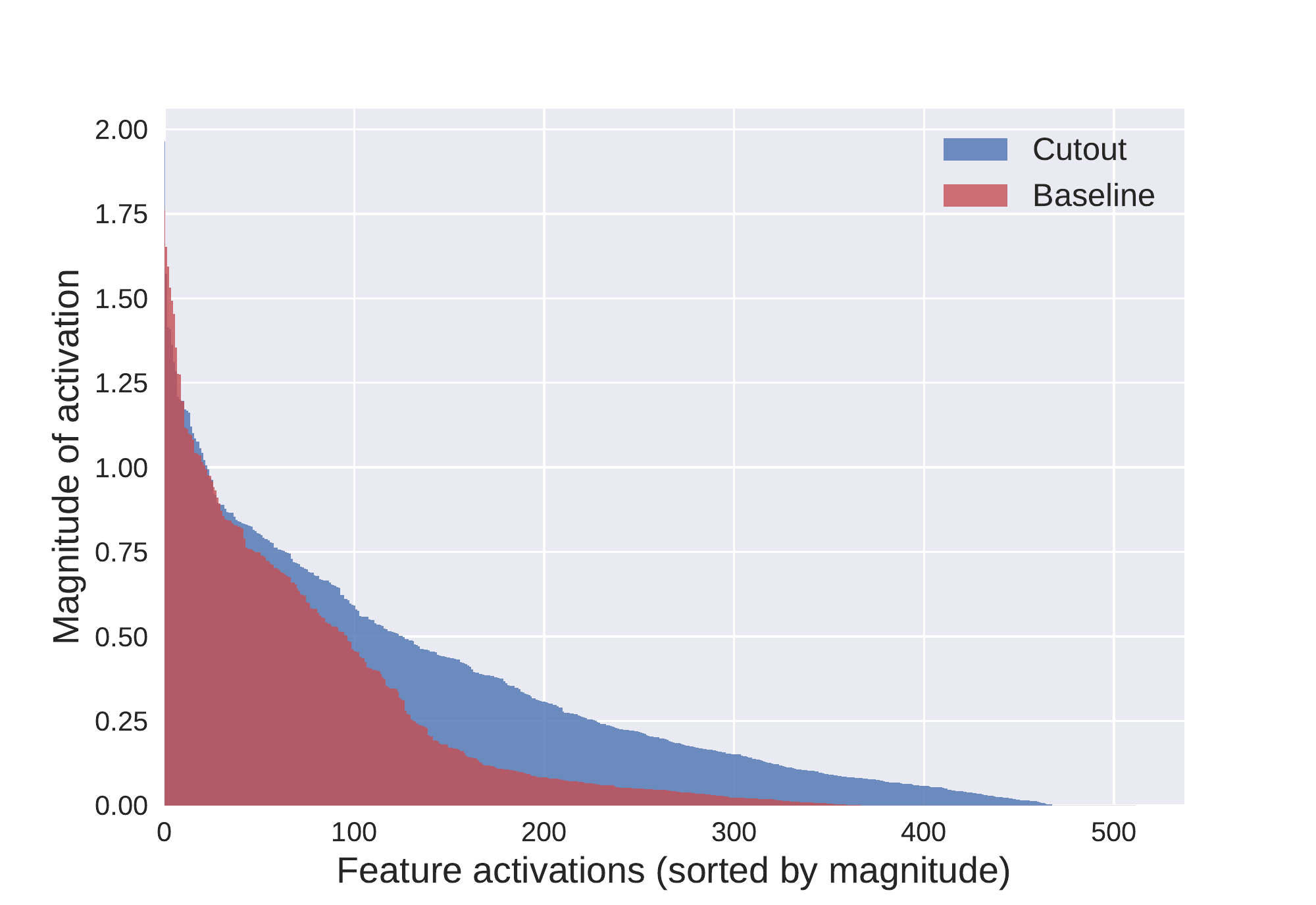}
 \subcaption{4th Residual Block}
\end{minipage}
\caption{Magnitude of feature activations, sorted by descending value. Each row represents a different test sample. A standard ResNet18 is compared with a ResNet18 trained with cutout at three different depths.}
\label{fig:act_hist}
\end{figure}

\end{document}